\documentclass[letterpaper]{article} 
\usepackage{aaai23}  
\usepackage{times}  
\usepackage{helvet}  
\usepackage{courier}  
\usepackage[hyphens]{url}  
\usepackage{graphicx} 
\usepackage{natbib}  
\usepackage{caption} 
\usepackage{algorithm}
\usepackage{algorithmic}

\usepackage{newfloat}
\usepackage{listings}
\DeclareCaptionStyle{ruled}{labelfont=normalfont,labelsep=colon,strut=off} 
\floatstyle{ruled}
\newfloat{listing}{tb}{lst}{}
\floatname{listing}{Listing}
\usepackage{amsfonts}
\usepackage{booktabs}
\usepackage{siunitx}
\usepackage{graphicx}
\usepackage{enumitem}
\usepackage{stix}
\usepackage{svg}

\newcommand\ROFT{RoFT}

\title{Real or Fake Text?: Investigating Human Ability to Detect Boundaries\\Between Human-Written and Machine-Generated Text}
\author {
    Liam Dugan\thanks{Equal Contribution},
    Daphne Ippolito$^*$,
    Arun Kirubarajan,
    Sherry Shi,
    Chris Callison-Burch
}
\affiliations {
    University of Pennsylvania\\
    \{ldugan, daphnei, kiruba, shershi, ccb\}@seas.upenn.edu
}

\begin{document}

\maketitle

\begin{abstract}
As text generated by large language models proliferates, it becomes vital to understand how humans engage with such text, and whether or not they are able to detect when the text they are reading did not originate with a human writer.
Prior work on human detection of generated text focuses on the case where an entire passage is either human-written or machine-generated.
In this paper, we study a more realistic setting where text begins as human-written and transitions to being generated by state-of-the-art neural language models.
We show that, while annotators often struggle at this task, there is substantial variance in annotator skill and that given
proper incentives, annotators can improve at this task over time.
Furthermore, we conduct a detailed comparison study and analyze how a variety of variables (model size, decoding strategy, fine-tuning, prompt genre, etc.) affect human detection performance.
Finally, we collect error annotations from our participants and use them to show that certain textual genres influence models to make different types of errors and that certain sentence-level features correlate highly with annotator selection.
We release the \ROFT{} dataset: a collection of over 21,000 human annotations paired with error classifications to encourage future work in human detection and evaluation of generated text. 
\end{abstract}

\section{Introduction}

Neural language models (LMs) are capable of generating increasingly natural-sounding text.
One growing worry is that bad actors may attempt to pass off automatically-generated text as genuine.
For example, \citet{zellersetal2019} discuss the dangers of machine-generated news articles, \citet{martens2019towards} document how easy it is to buy fake app store reviews, and 
\citet{weidinger2021ethical} chronicles how LMs can potentially be used to spread misinformation, fraud, and other harmful text. 
These harms will inevitably become more and more prevalent as language models become better and cheaper to deploy.
Thus, it is important to answer the question: just how susceptible are humans to being duped by machine-generated text?

Existing studies of the ability of humans to detect generated text have focused on the binary question of whether or not a provided document contains any generated text at all
\citep{ippolito-etal-2020-automatic,gehrmann2019gltr, clark2021all}.
In this work, we instead frame detection as a boundary-detection task: given a document that starts off as human-written and at some point transitions to machine-generated, can annotators detect the transition point?
The boundary detection setting is more informative than the classification setting because it better aligns with how LMs are used to generate text in practice---in typical usage, a generative system is provided with a prompt and asked to produce a continuation.
By measuring human skill at the boundary detection task, we can make progress toward quantifying the risks associated with LMs; while simultaneously evaluating the performance of different generative systems.

\begin{figure}[tb]
    \center
    \framebox{\includegraphics[scale=0.19]{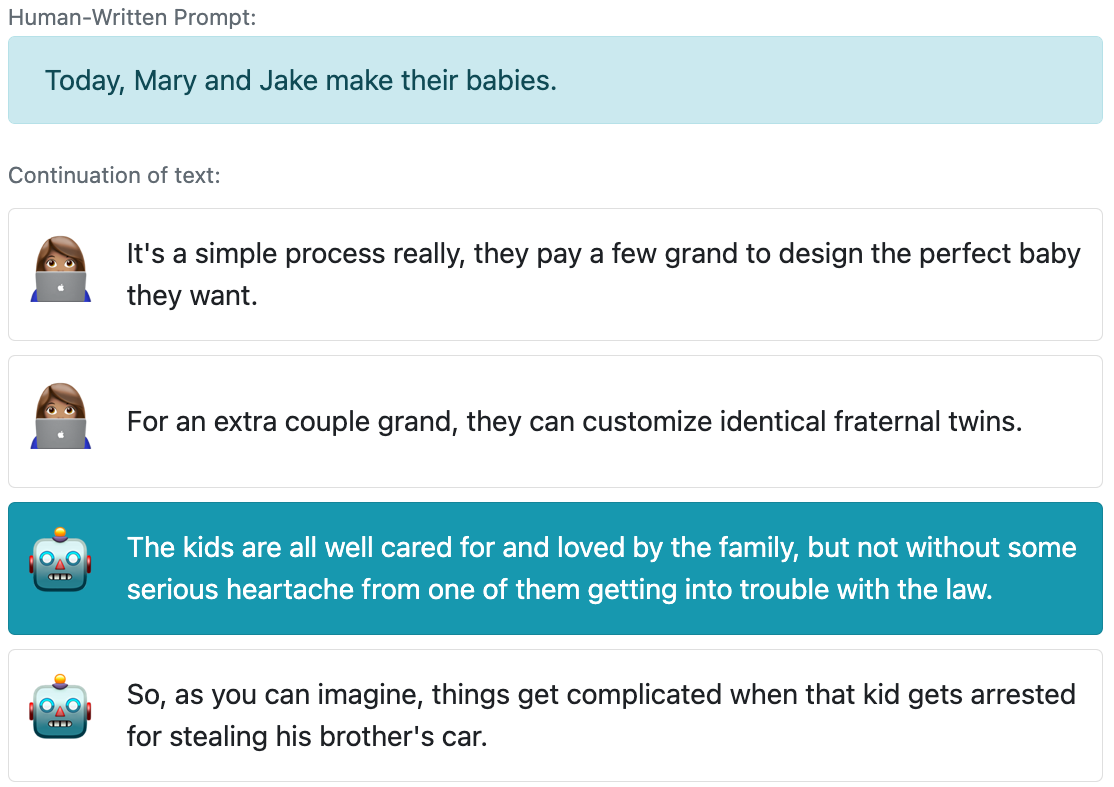}}
    \caption{In the boundary detection task, players see one sentence at a time from a passage and try to guess when the passage transitions from human-written to machine-generated.}
    \label{fig:pg1}
\end{figure}

We collect annotations using \ROFT{}, the website introduced in \citet{roft} which poses the boundary detection task as a game.
In each game round, players are shown one sentence at a time and earn points for guessing close to the true boundary (Figure \ref{fig:pg1}).
They are also asked to select a reason for why they made their decision. In our analysis of these annotations, we find that players vary substantially in their detection ability and that factors such as the amount of time taken to complete a game round and total number of game rounds played correlate with success.

We discuss how various factors such as the genre of prompt (news, stories, etc.), the size of the model, and decoding strategy used affect generation quality. Finally, we examine the trends and errors which distinguish real from generated text and look at whether our players could pick up on these trends.
In addition to producing valuable data for analyzing detectability, our study serves as the first large-scale attempt at using a gamified platform to analyze the detectability of generated text. Such a platform is easily extensible to support answering additional research questions in the future.
All generations and annotations used in this paper are made publicly available to encourage further study of the detectability of machine-generated text\footnote{https://github.com/liamdugan/human-detection}.

\section{Related Work}
Previous research on understanding the ability of humans to detect machine generated text has mostly posed the task as a classification task---given a text example that is either entirely human-written or entirely machine-generated (aside from an initial prompt), annotators must predict whether it is human-written or machine-generated.
On this task, \citet{ippolito-etal-2020-automatic} reported that trained evaluators were able to achieve an accuracy of at best 71.4\%, using generations from GPT-2 Large \citep{radford2019language}.
In a follow-up study, \citet{clark2021all} demonstrated that annotators are able to distinguish GPT-2 XL generations with at best 62\% accuracy, but they perform no better than random chance on GPT-3 \citep{brownetal2020} outputs. 

Even after training evaluators to improve their detection abilities, detection accuracy on GPT-3 was only able to converge to around 55\% \citep{clark2021all}.
A study by \citet{brownetal2020} reported similarly low performance (52\%) on the detection of machine-generated news articles. Most recently, \citet{Ethayarajh2022} report that annotators rate GPT-3 outputs as significantly more human-like than human-written text itself, suggesting a detection accuracy of \emph{worse} than random chance, underscoring a need to re-think our human evaluations from both a metrics perspective and a task perspective.

In this vein, another related area of research is asking annotators to explain why they think generated text is generated.
\citet{he-etal-2021-tgea} created a dataset of generated text annotated with the errors that humans found in it, and \citet{dou2021scarecrow} proposed an error annotation schema for generated text.
The errors we allow players to report in our experiments were inspired by these schemata.

Our work aims to address this need to diversify our evaluation task by using the RoFT detection game format introduced by \citet{roft} to analyze human detection performance across a variety of genres and generative systems.
A similar detection task, where annotators guess whether turns in a conversation were generated, was posed by \citet{deriu-etal-2020-spot} as a way to evaluate dialog systems, but it has yet to be applied to language models more generally. We demonstrate the feasibility of applying such an evaluation framework to more general language models.

\begin{table*}[tb]
\center
\small
\begin{tabular}{l|r|cc|r|c|c}
\toprule
 & \# & \multicolumn{2}{c|}{\# Annotations} & {Avg} & & \\
Genre & Avail & Raw & Final & Ann/Cont. & Generation Sources & Decoding Strategies \\
\midrule
News & 1,838 & 7,806 & 4,488 & 2.97 &
\begin{minipage}{.25\textwidth}
      \includegraphics[height=1em]{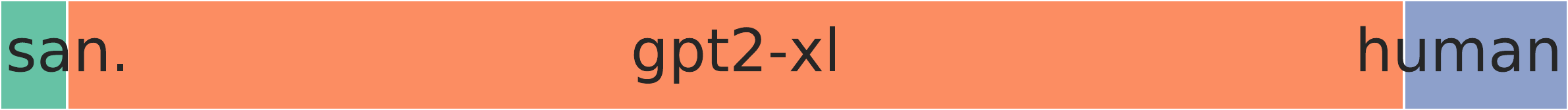}
\end{minipage}
&
\begin{minipage}{.20\textwidth}
      \includegraphics[height=1em]{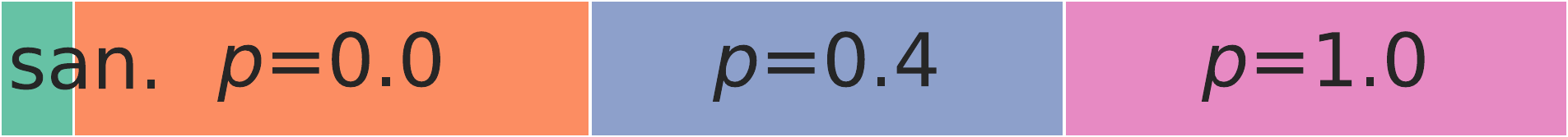}
\end{minipage}
\\
Stories & 9,864 & 8,007 & 4,614 & 2.53 &
\begin{minipage}{.25\textwidth}
      \includegraphics[height=1em]{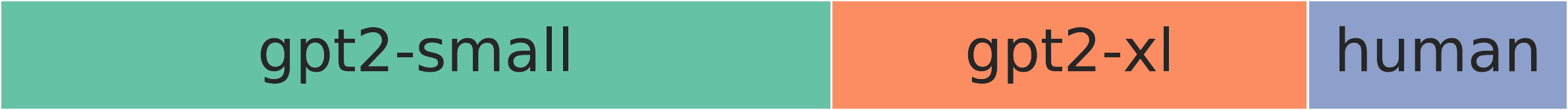}
\end{minipage}
&
\begin{minipage}{.20\textwidth}
      \includegraphics[height=1em]{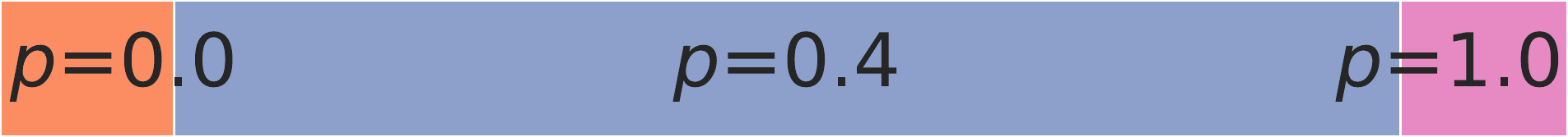}
\end{minipage}
\\
Recipes & 7,258 & 17,978 & 7,709 & 2.13 &
\begin{minipage}{.25\textwidth}
      \includegraphics[height=1em]{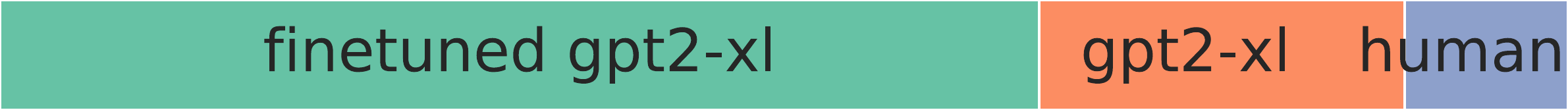}
\end{minipage}
&
\begin{minipage}{.20\textwidth}
      \includegraphics[height=1em]{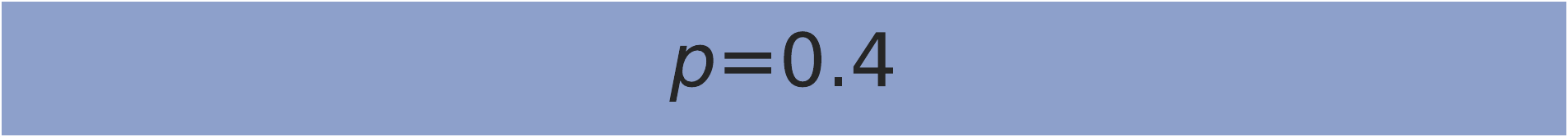}
\end{minipage}
\\
Speeches & 297 & 8,374 & 4,835 & 16.28 &
\begin{minipage}{.25\textwidth}
      \includegraphics[height=1em]{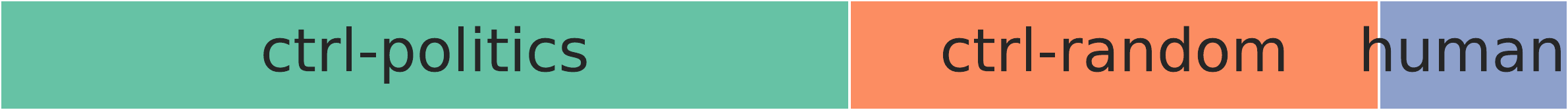}
\end{minipage}
&
\begin{minipage}{.20\textwidth}
      \includegraphics[height=1em]{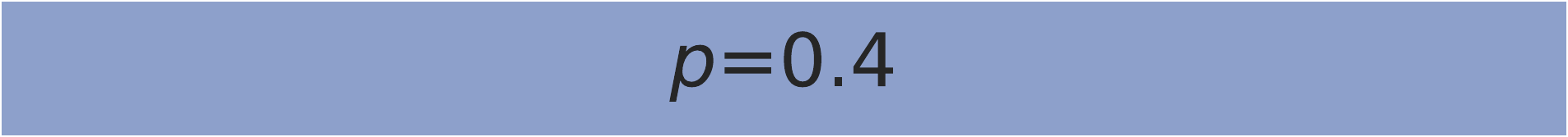}
\end{minipage}
\\
\bottomrule
\end{tabular}
\caption{Statistics on the game rounds available and annotations collected.
Players were asked to play 50 rounds in an assigned genre, after which they could choose any genre.
Recipes was the most popular genre.
}
\label{tab:dataset_stats}
\end{table*}

\section{Experimental Design}
\subsection{The Real or Fake Text Game}
Our study uses data collected through the ``Real or Fake Text'' (\ROFT{}) annotation platform \citep{roft}.
\ROFT{} is a turn-based game where a player first selects a domain of text (news articles, recipes, short stories, or speeches).
The player then plays a series of game rounds.
In each round, the player is shown a starting sentence which they are told comes from a real human-written document.
They are then shown subsequent sentences, one at a time.
Each subsequent sentence may be the true continuation of the document, or it may be text generated by a language model.
Once the sentences transition to being machine-generated, they will stay so for the rest of the 10-sentence passage.

After being shown each sentence, the player must decide whether they think that sentence was machine-generated or human-written.
If the user selects ``human-written,'' another sentence is displayed.
If the player selects ``machine-generated,'' the game round ends and the true author (machine or human) for each sentence is revealed.
Before submitting their selection, the player is able to select a reason to explain their choice of sentence.
Thus, the player's goal in \ROFT{} is to correctly identify the sentence at which a passage transitions from being human written to being generated by a language model.
We claim that this task formulation is more faithful to how generated text appears in real world scenarios, since generating with a prompt is the standard way to achieve controllability, and malicious actors will not reveal what portion of a generation is the human-written prompt.

\subsection{Datasets}
For our study we sampled prompts from four diverse genres of text. For each genre, we selected a corpus of text from the genre, sampled documents from the corpus, sentence-segmented them, and filtered out all documents with less than ten sentences. 
Then, for each document, we randomly select one of the ten sentences to be the end of the prompt and replace all following sentences with a machine generated continuation. This results in an even distribution over prompt lengths, with approximately 10\% of examples being fully human-written. Our four genres of prompts are as follows:

\paragraph{News Articles.}
Documents were drawn from the New York Times Annotated Corpus \citep{sandhaus2008new}, which contains 1.8 million articles published by the Times between 1987 and 2007.
Our hypothesis was that this domain would be challenging for models since news requires factual accuracy, which state of the art models have been shown to struggle with \cite{nakano2021webgpt, lin2022truthfulqa}. 

\paragraph{Presidential Speeches.}
Documents were drawn from the presidential speech corpus \citep{brown2016cops}, which contains 963 speeches given by presidents of the United States, with dates ranging from 1789 to 2015.
Our hypothesis was that the sort of first-person rhetoric found in these speeches would be easy for models to impersonate since political speech and first-person speech are plentiful in web-based training data.

\paragraph{Stories.}
Fictional stories were drawn from the Reddit Writing Prompts dataset \citep{fan2018hierarchical}, a corpus of amateur short stories scraped from the r/WritingPrompts subreddit\footnote{https://www.reddit.com/r/WritingPrompts/}. 
We hypothesized that this domain would be easy for models since the writing quality of the stories is not especially high (which lowers the bar for the model generation quality), and factuality is not as important in a fictional domain.

\paragraph{Recipes.}
Recipes were extracted from the Recipe1M+ dataset \cite{marin2019learning}.
Recipes were parsed slightly differently than the other domains.
We set the first ``sentence'' of each document as the name of the recipe and the ingredient list, and each subsequent ``sentence'' was a step in the recipe. Some recipe steps were more than one sentence. We hypothesized that this dataset would be difficult for models due to the closed-ended and structural nature of the task and the reliance on common sense. 

\subsection{Awarding Points}
\label{sec:points}
In each game round, the player is awarded points based on how close their selection was to the true boundary sentence (i.e. the first machine-generated sentence in the passage).
Players were awarded 5 points for correctly choosing the boundary sentence and $\max(5-n, 0)$ points for a guess $n$ sentences after the boundary.
Players were not awarded points for guessing a sentence before the boundary.
Players were able to see how many points they earned in each category on their profile page and compare their performance with fellow players on the leaderboard page.
In the Findings section (Section \ref{sec:findings}), we report mean score earned as the predominant evaluation metric. 
This metric has high correlation with other more standard metrics (see Appendix \ref{app:metric} for more detail).

Some students did discover that by always choosing the last sentence, they could game the system to earn points without putting in effort.
We accounted for this tendency and filtered out all suspicious annotations (see Appendix \ref{app:filtering}).

\subsection{Players}
\label{sec:players}
Players were recruited from two sections of an Artificial Intelligence course for graduate students and senior undergraduates at the University of Pennsylvania.
Each participant was randomly assigned a single genre for their first 50 annotations, after which they were allowed to choose between genres.
All data used in our study is fully anonymized and only collected from students who explicitly consented to having their annotations used for research purposes. Additionally, the ethics of this study were reviewed and approved by the University of Pennsylvania's Institutional Review Board.

We separated our participants into two groups. Group A was our control group. They were asked to play 30 minutes of the \ROFT{} annotation game and received 2 points of class extra credit as compensation regardless of their score.
The second section (Group B) was explicitly told they would be awarded $\min(p/250, 2)$ points of extra credit toward their final grade, where $p$ was the total number of points the student earned across all rounds played.
While Group A was only given basic instructions on how to play the game, Group B was given a detailed guide for how to identify generated text. 
Statistics on the number of annotations collected from each group can be found in the Appendix.

We note that university students taking an advanced artificial intelligence course are not reflective of the global population of English speakers, and the results presented in this paper may not reflect the general population's ability to detect machine-generated text. However, we believe our set of students are sufficient for a preliminary study and we would like to see broader representation in future work.

\subsection{Continuation Sources}
\label{sec:experiments}
One of the main goals of our study is to investigate how specific model attributes (size, sampling strategy, etc.) affect our players' ability to detect generated text. In order to do this we structure our continuation sources as follows: We first decide on a base model configuration, then for each experiment we vary exactly one aspect of that base model and observe how human performance changes. We generally use only one (sometimes two) genres per comparison due to budget constraints but encourage future work to address this limitation.

For our base continuation model we use pre-trained GPT-2 XL with nucleus sampling parameter of $p=0.4$ \citep{holtzmanetal2020} and repetition penalty of $1.2$ \citep{keskar2019ctrl}. We generate continuations using this model on News, Stories, and Recipes to serve as our base for comparison. As an additional sanity check on annotator skill, we also include 100 game rounds in the News domain where instead of transitioning to an LM-generated continuation, the passage transitions to a completely different news article selected at random.
We expected these game rounds to be trivial for players.

For our model comparisons, we first generate continuations on Stories with GPT-2 small in order to investigate how model size affects detectability. Then, we investigate whether or not noisy text is easier to detect by generating continuations on News and Stories with GPT-2 XL using $p=0.0$ (argmax sampling) and $p=1.0$ (random sampling). In our third comparison, we fine-tune GPT-2 XL on Recipes and compare the output to our pre-trained base model. Finally, for the last comparison we look at the effects of topic control codes by generating continuations on Speeches using the CTRL model \cite{keskar2019ctrl} rather than GPT-2.

CTRL is a 1.6B parameter LM that allows users to pass in one of 50 pre-defined control codes to condition the model to generate in a particular style. For half of the generations, we used the ``[Politics]'' control code while for the other half we randomly selected a control code each time. We hypothesized that presidential speeches would be a good domain for this comparison because speeches have a very distinctive formal style and are typically very semantically similar.

We report the results of our main study of detectability in Section \ref{sec:findings}. We report the results of our comparison experiments in Section \ref{sec:exploration}. Finally, we report extra \ROFT{}-Specific analysis on topics such as model errors in Section \ref{sec:insights}.

\subsection{Final Annotations Used for Analysis}
In total, we collected 42,165 annotations over 7,895 different game rounds. After filtering for an even distribution over prompt length and removing players who exploited vulnerabilities (see Appendix \ref{app:gen_filter}), we ended up with a final dataset of 21,646 annotations over 7,257 continuations.

Table \ref{tab:dataset_stats} gives a detailed breakdown of the dataset across genres and generation systems. For News, Stories, and Recipes, we had on average 2 players per continuation, while for Speeches, a smaller dataset, we had 16.

\section{Main Findings}
\label{sec:findings}
In this section we report the findings of our main detectability study. We analyze the accuracy of our players, their agreement, their improvement over time, the distribution of skill across players, and if said skill varies across genre.
Error bars on all figures are 95\% confidence intervals and the exact values and confidence intervals for all figures can be found in Appendix \ref{app:graph_values}.

\paragraph{Can humans detect generated text?}
We found that players were significantly better than random chance at the boundary detection task, correctly selecting the boundary sentence 23.4\% of the time (chance being 10\%).
For rounds with at least one generated sentence, players selected a generated sentence as the boundary sentence 72.3\% of the time.

The average number of points (\S\ref{sec:points}) 
received per round by our players was 2.08, also well above random chance (1.31 points). Additionally, out of the 214 annotations we collected for our ``sanity check'' baseline, the mean score was 2.75, significantly higher than any of the true LM-backed systems (but still far from a perfect 5.00). 
For the remaining analyses, we will use average points earned per round (``mean score'') as the primary measure of detection ability instead of accuracy.
This measure correlates with accuracy as well as other metrics that one might consider using (see Appendix \ref{app:metric}).

To measure inter-annotator agreement we use the Krippendorff's alpha co-efficient.
This statistic measures how much disagreement there is between players compared to the amount one would expect by chance.
Two players are considered to have agreed if they both guessed ``machine-generated'' on any sentence on or after the true boundary or if they both guessed the entire passage was human-written.
Over all annotations, we found $\alpha$=0.25, indicating only slight agreement amongst most players. However, among our top 10\% of
players (measured by mean score), there was high agreement, with $\alpha$=0.40, suggesting that good players made similar errors.

\begin{figure}[tb]
    \center
    \hspace*{-0.5cm} 
    \includegraphics[scale=0.5]{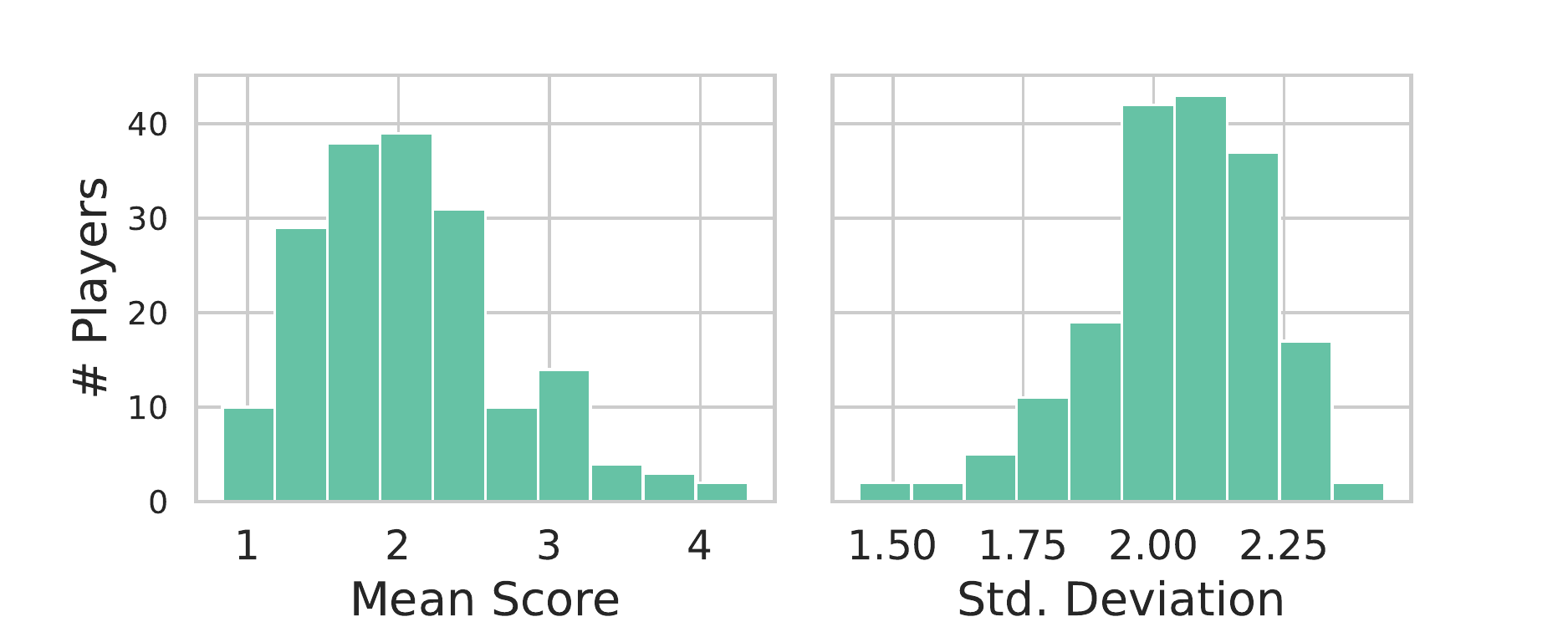}
    \caption{Histogram of mean score and standard deviation of score among players who completed at least 20 rounds. We see large gaps in skill between players, with some having significantly higher mean score and lower variance than others.}
    \label{fig:mean_var}
\end{figure}

\paragraph{How much does player ability vary?}
We observed a large variance in the skill of individual players. In Figure \ref{fig:mean_var} we report the distribution of mean scores and standard deviations of all players that completed more than 20 rounds. We
see that some players
have significantly higher mean score than others and we find that higher mean score also tends to coincide with lower standard deviation. Among players with mean score over 3.8, the average standard deviation was 1.84.

\begin{figure}[tb]
    \center
    \hspace*{-0.25cm} 
    \includegraphics[scale=0.5]{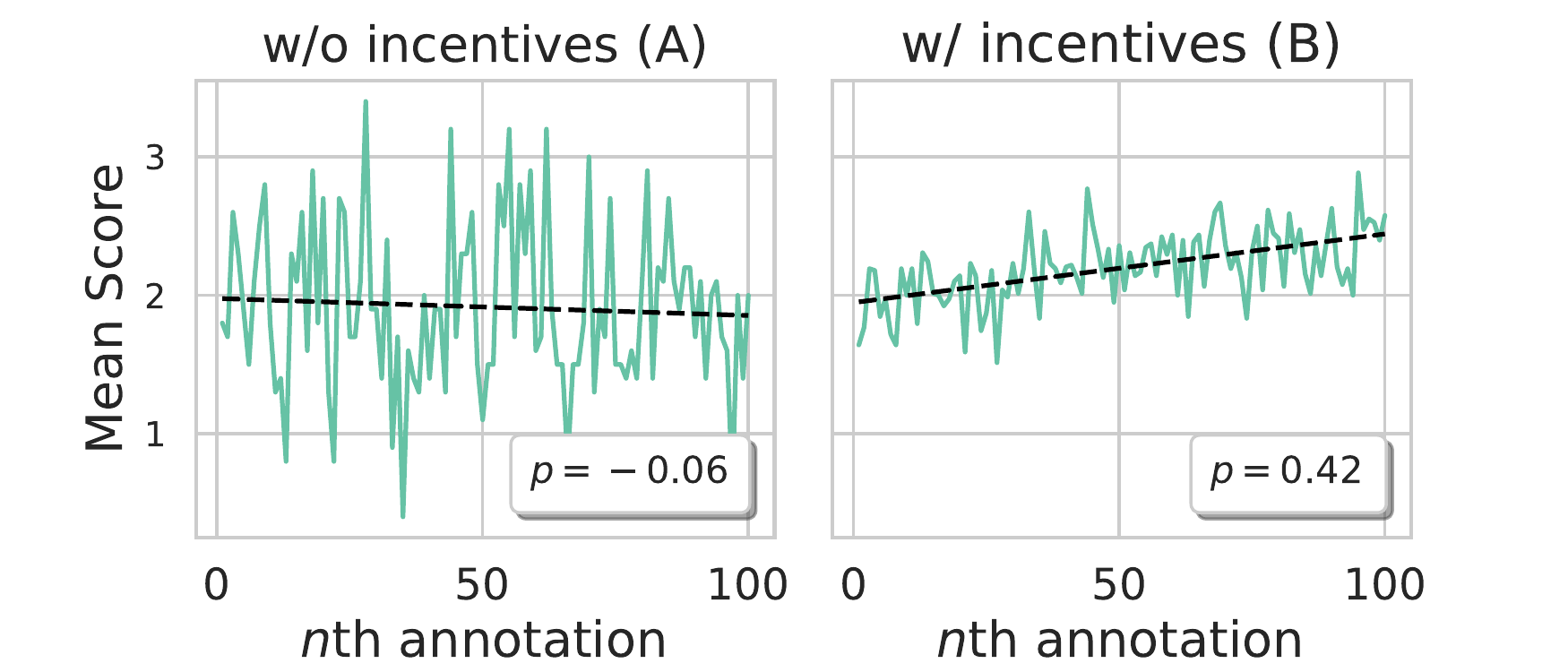}
    \caption{Performance over time for the two player groups (\S\ref{sec:players}). Players in Group B, who were given extra instruction and incentives, improved over time ($\rho=0.42$) while those in Group A did not ($\rho=-0.06$).}
    \label{fig:skill_over_time}
\end{figure}

We also found that under the right conditions, players can exhibit improvement over time.
In Figure \ref{fig:skill_over_time} we report the performance over time across our two player groups. While we saw no correlation between number of rounds played and player score for our control (Group A), Group B, who were given extra credit proportional to their game score and extra instruction, did show improvement along with lower variance over time.

Finally, among all questions asked on our exit survey, we found that the most predictive feature for annotator mean score was whether or not they reported that they had read our help guide\footnote{http://github.com/liamdugan/human-detection/data/guide.pdf}. This guide contained a taxonomy of model errors, a set of annotated examples, and other detection tips and common pitfalls. This finding, coupled with the previous findings on improvement over time and high skill variance, suggest that detection is a skill and that annotators can be trained to do well on the detection task.

\begin{figure}[tb]
    \center
    \hspace*{-0.3cm}
    \includegraphics[width=0.95\linewidth]{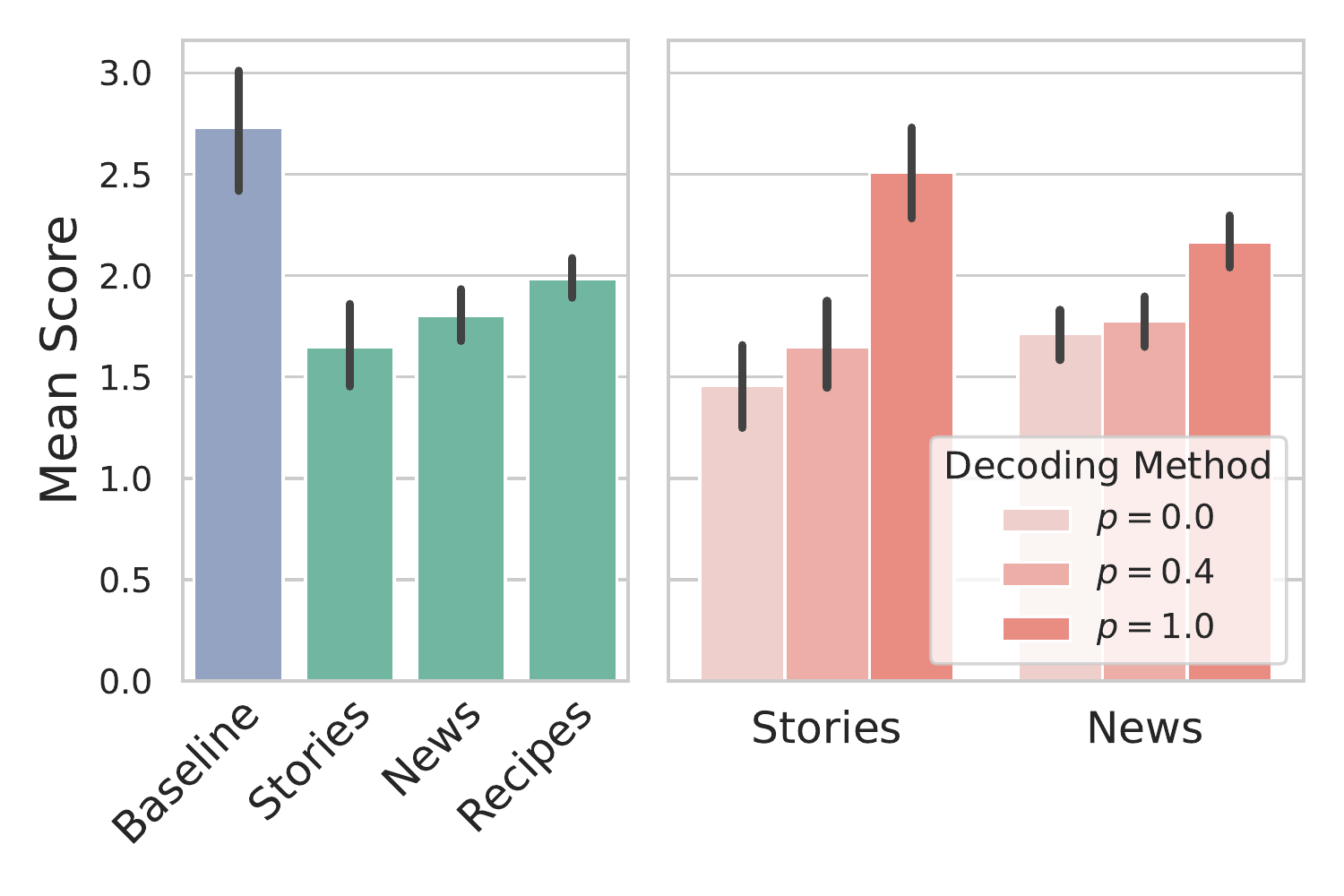}
    \caption{\textbf{(left)} Comparison of mean player score across different genres with GPT-2 XL $p$=0.4 against our ``sanity-check'' baseline (\S\ref{sec:experiments}). \textbf{(right)} Comparison of mean player score across values of $p$ for nucleus sampling (GPT-2 XL).}
    \label{fig:genre_top_p}
\end{figure}

\begin{figure}[tb]
    \center
    \hspace*{-0.5cm}
    \includegraphics[width=0.8\linewidth]{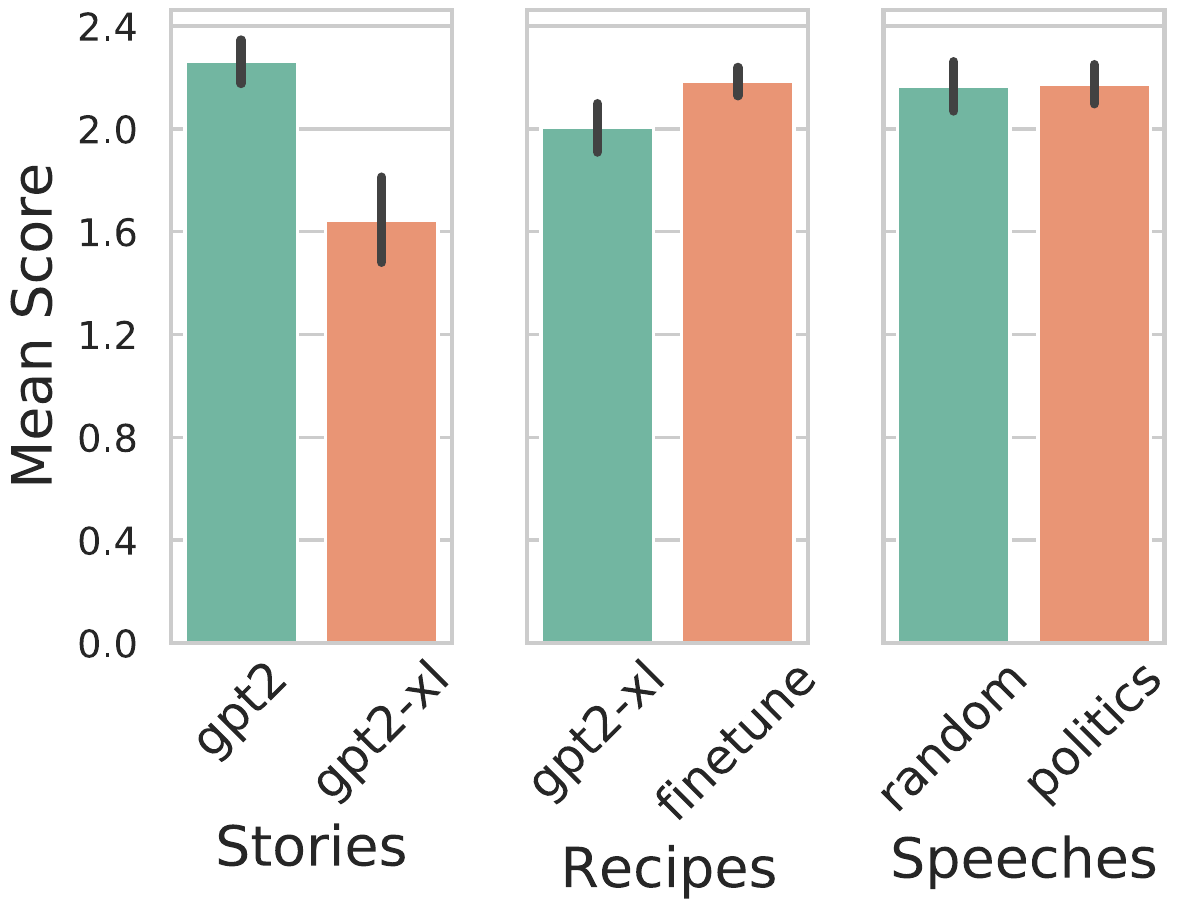}
    \caption{\textbf{(left)} For Stories, as model size increases (using $p$=0.4), detection becomes harder. \textbf{(middle)} For Recipes, extra finetuning does not significantly impact detectability. \textbf{(right)} For Speeches, using a ``[Politics]'' control has no impact on detectability.}
    \label{fig:model_size_finetuning}
\end{figure}

\begin{figure*}[tb]
    \center
    \includegraphics[width=0.9\linewidth]{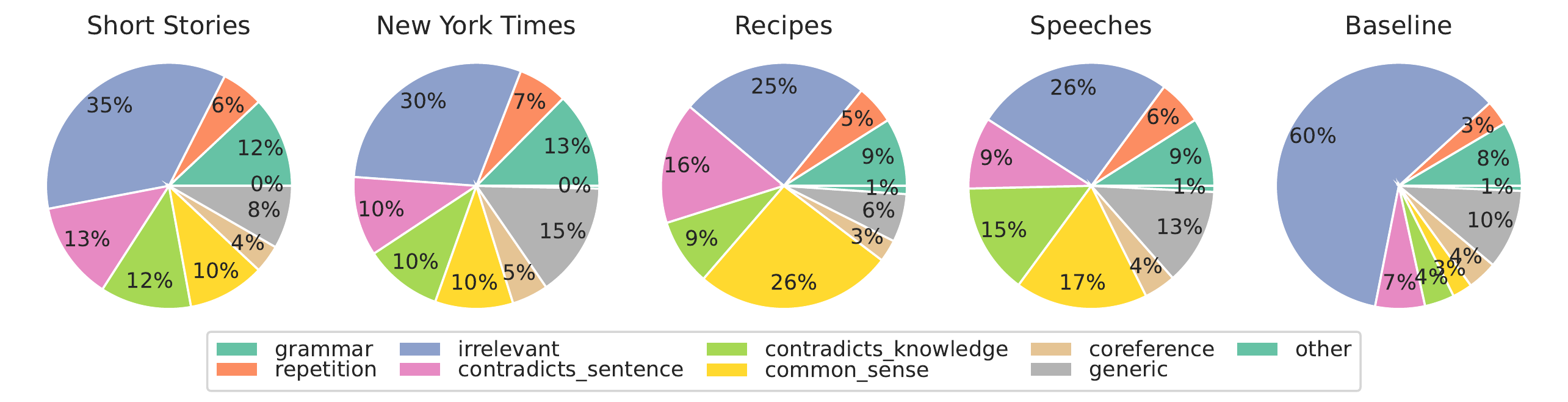}
    \caption{The reasons given by our players as to why they thought a given sentence was machine generated. Continuations for Stories, News, and Recipes were generated by GPT-2 XL with $p=0.4$. Baseline refers to our sanity check baseline (\S\ref{sec:experiments}). We see that GPT-2 XL tends to make more ``common\_sense'' errors on recipes, more ``irrelevant'' errors on stories, and more ``generic'' errors on news. This is consistent with our intuition about what makes each of these domains challenging.}
    \label{fig:reasons_per_domain}
\end{figure*}

\begin{figure}[tb]
    \center
    \includegraphics[width=0.9\linewidth]{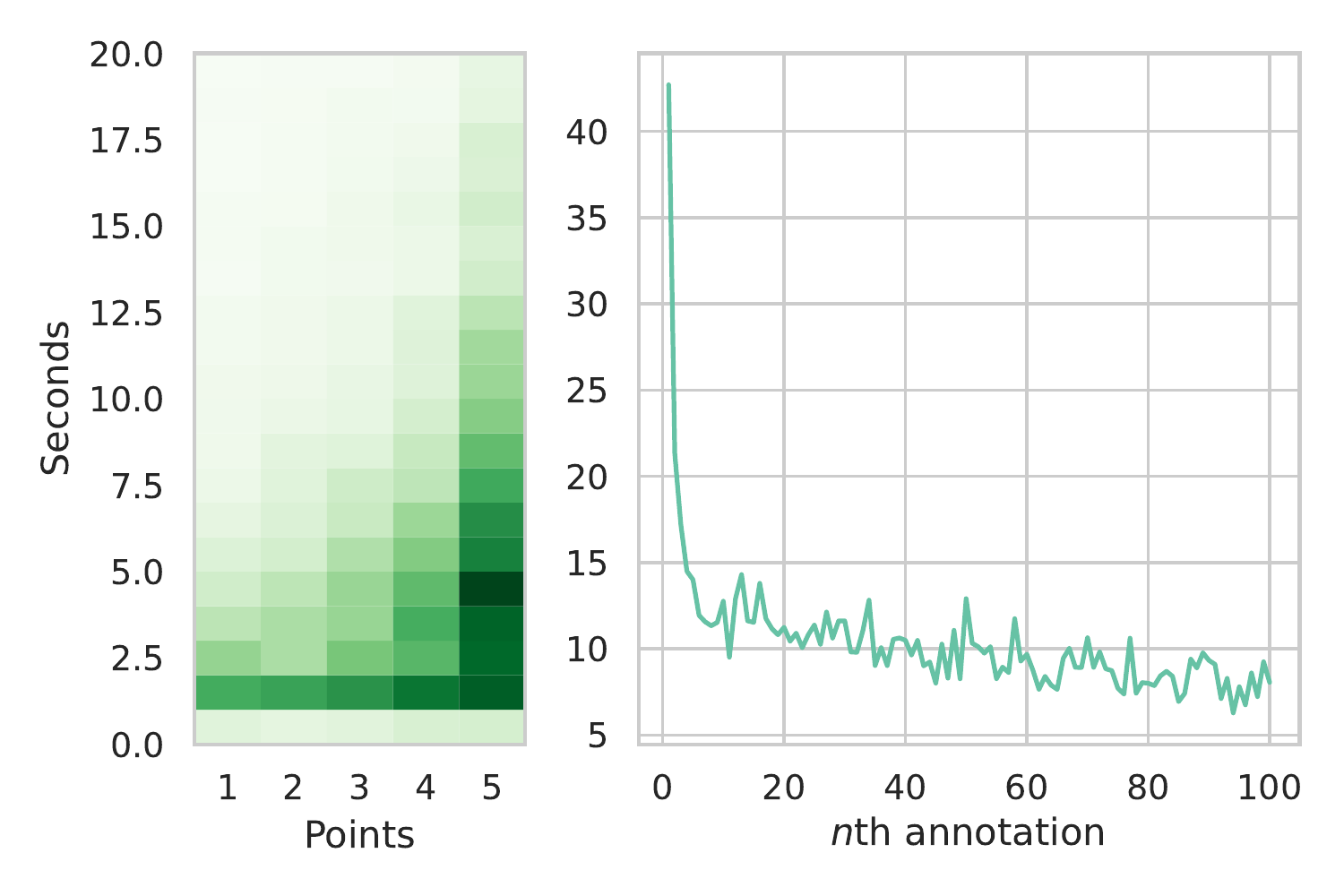}
    \caption{\textbf{(left)} Histogram showing the time taken per annotation and the score received.
    Scores tended to be higher when more time was taken.
    \textbf{(right)} Plot of time taken per annotation over time. We see that players get faster at the task over time.}
    \label{fig:time_tracking}
\end{figure}

\paragraph{Are some genres easier to detect?}
We found that certain genres were slightly easier than others. In particular, generated recipes were easier to detect than generated stories or news articles.
Figure \ref{fig:genre_top_p} (left) shows a comparison of mean player score across each genre.

For Recipes, we hypothesize that detection was made easier by the fact that the first human-written ``sentence'' in each game round was a semi-structured ingredients list, making it easy for players to check for contradictions---a step saying to mix in cream is probably generated if there is no cream ingredient.
In addition, recipes often assume implicit unwritten knowledge which language models struggle to get right---a step saying to crack eggs cannot follow a step saying to whisk the eggs. 

Indeed, if we look at the reasons given by our players as to why they selected certain recipes as generated (Figure \ref{fig:reasons_per_domain}), we see that continuations in the Recipes domain contain a much larger percentage of ``common\_sense'' errors (26\%) than those in the News (10\%) or Stories (10\%) domain.

\section{Model Comparison Findings}
\label{sec:exploration}
In this section we report the results of our comparison experiments. These are one-off comparisons that investigate the effect of one particular variable on detection accuracy. Similar to Section \ref{sec:findings}, error bars on all figures are 95\% confidence intervals and the exact values and confidence intervals for all figures can be found in Appendix \ref{app:graph_values}.

\paragraph{Does model size affect detection performance?}
Previous work has shown that language model performance scales with number of parameters \citep{kaplan2020scaling, hoffmann2022training}, so we expected players to be worse at detecting generations from larger models.
Indeed, we found that players scored significantly higher when generations came from GPT-2 small (117M parameters) than when they came from GPT-2 XL (1.5B parameters).\footnote{In an additional round of annotation, we also experimented with continuations from GPT-3 Davinci (175B parameters) but could not draw any statistically significant conclusions. We have included the results from that round of annotation in Appendix \ref{app:gpt3}.} In Figure \ref{fig:model_size_finetuning} (left) we report the difference in mean player score between GPT-2 small and GPT-2 XL. The difference observed here is the most significant difference observed across all variables tested, reaffirming the correlation between scale and language model performance. 

\paragraph{Are diverse generations easier to detect?}
Choice of decoding strategy is known to have significant impact on text quality \citep{zhang-etal-2021-trading} and detectability \citep{ippolito-etal-2020-automatic}.
Choosing a lower value of $p$ when generating with a nucleus sampling \citep{holtzmanetal2020} decoding strategy produces less diverse but also less noisy text than choosing a higher value of $p$. In Figure \ref{fig:genre_top_p} (right) we report our findings and see that players were significantly better at $p$=1.0 (pure random sampling) than the lower values, 
validating claims from earlier papers that LMs struggle to generate high-quality text with similar diversity to human-written text.

\begin{figure*}[tb]
    \center
    \includegraphics[width=0.9\linewidth]{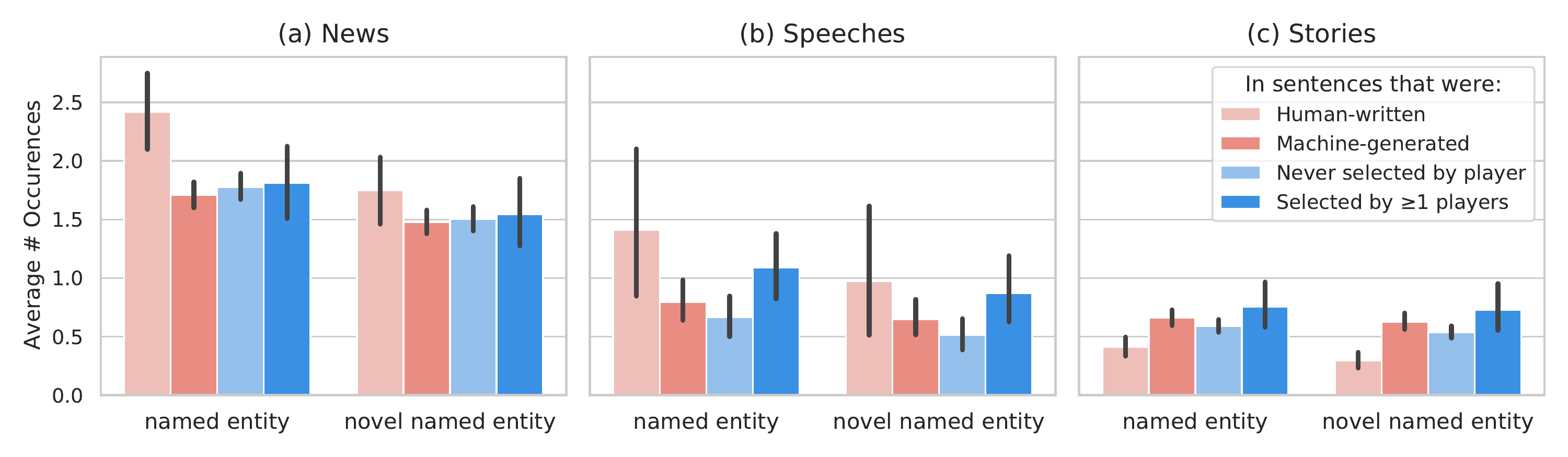}
    \caption{We see that human sentences tended to have a different number of named entities than generated sentences.
    Players picked up on the correct trend in Stories, but not in News or Speeches.}
    \label{fig:sentence_stats}
\end{figure*}

\paragraph{Does finetuning affect detectability?}
We had expected that finetuning on in-domain text would result in a model that was better able to fool humans.
We report our results in Figure \ref{fig:model_size_finetuning} (middle) and see that, counter to expectations, there was a small increase in player detection ability when generations came from GPT-2 finetuned on recipes compared with generations from pre-trained GPT-2. This is despite the fact that the finetuned model had close to half the perplexity of the pre-trained model on a held out test set of 50,000 recipes (4.781 vs. 8.979).
While we can only speculate as to the amount of recipe knowledge present in the pre-trained model (GPT-2's training data is not publicly available), it is possible the pre-trained model already contained enough understanding of recipe-like text that it was not critical to do the extra-finetuning. 
We speculate that finetuning may have had more impact in a specialized or jargon-laden domain (e.g. legal, medical).

\paragraph{Do control codes affect detectability?}
CTRL is a 1.6B parameter LM trained with controllability in mind.
At inference time, one can pass in one of 50 pre-defined control codes, such as ``[Politics]'' or ``[Horror]'', to condition the model to generate in a particular style.
We investigated the efficacy of these control codes on the genre of presidential speeches by using ``[Politics]'' for half of the generations and randomly selecting a control code for the remaining half. We decided to use the presidential speeches genre for this experiment due to its unique and distinctive style.
We report our results in Figure \ref{fig:model_size_finetuning} (right) and find that use of the politics control code did not significantly affect players' ability to distinguish real from fake text.
This is not to say that control codes do not affect generation; however, it does suggest that the cues used by players to detect generations may not be related to stylistic details, at least not within the genre of political speeches.
Further work is needed to investigate whether control codes could have influenced detectability in other more specialized domains (e.g. legal, medical).

\section{\ROFT{}-Specific Insights}
\label{sec:insights}
In this section we use specific capabilities of the \ROFT{} platform such as time-tracking and sentence-level error annotation to investigate additional research questions about how and why humans select certain sentences as generated.

\paragraph{How much time did game rounds take?}
To understand how much time game rounds took, we logged the amount of seconds players spent on each sentence decision.
We controlled for instances of players leaving a game open mid-annotation by applying $\min(120, t)$ to all recorded times $t$.
We computed total time per annotation by summing the times for each sentence-level decision. In Figure \ref{fig:time_tracking} we report the time taken per annotation (left) and the time taken per annotation over time (right).
Unsurprisingly, when players took longer on annotations, we found that they ended up receiving more points. We also found that players gradually got faster over time.
Interestingly, while one might expect longer sentences to take more time to read and make decisions on, we found no correlation between time taken and length of sentence ($\rho$=-0.10). This indicates that players take time to think about the task beyond just reading the sentence.

\paragraph{What errors do humans look for when detecting generated text?}
\label{sec:reasons}
Each time a player specified a sentence was machine-generated, they were asked to specify why they made this decision, selecting from a set of pre-defined options or otherwise writing down a custom reason. Figure \ref{fig:reasons_per_domain} shows the distribution of these reasons across textual genres. For the Recipes domain we see a much higher proportion of ``common\_sense,'' errors, in News we see models generating much more ``generic'' text, and in Stories we see more ``irrelevant'' text. 

Table \ref{tab:reasons} shows, for each reason, the ``reliability'' of the reason, i.e. the average number of points earned on rounds where that reason was given.
Like \citet{clark2021all}, we see that conditioning on bad grammar is by far the least reliable way to detect generated text with a mean score of 1.78.
Interestingly, the three most common reasons given (``common\_sense,'' ``irrelevant,'' and ``contradicts\_sentence'') were also the three most reliable, indicating that improving these attributes will lead to the biggest improvements in generation performance (and ability to fool humans). 

\begin{table}[tb]
\small
\center
\begin{tabular}{lcc} \toprule
    {Reason} & {$n$} & {Mean Score}\\ \midrule
    {common\_sense} & {2,432} & {2.566 $\pm$ 0.086} \\
    {irrelevant} & {4,259} & {2.530 $\pm$ 0.064}  \\
    {contradicts\_sentence} & {1,606} & {2.527 $\pm$ 0.105}  \\ 
    {contradicts\_knowledge} & {1,411} & {2.262 $\pm$ 0.111}   \\
    {coreference} & {542} & {2.249 $\pm$ 0.176} \\
    {repetition} & {728} & {2.128 $\pm$ 0.154} \\
    {other} & {75} & {2.040 $\pm$ 0.483} \\
    {generic} & {1,546} & {1.920 $\pm$ 0.101} \\
    {grammar} & {1,539} & {1.780 $\pm$ 0.105} \\\bottomrule
\end{tabular}
\caption{
The number of times each reason was given for text being machine-generated, and the mean score over those annotations.
We see that when players select reasons like ``grammar'' or ``generic,'' they are much less likely to be correct than when selecting ``common\_sense'' or ``irrelevant.''}
\label{tab:reasons}
\end{table}

\paragraph{What sentence-level features could be used to detect generated text?}
It has been well-studied how generated text differs in basic, measurable ways from human-written text, often due to the choice of decoding strategy.
Given our task format, we wanted to measure how sentence length, part-of-speech distribution, and presence of novel named entities differed between the generated and human-written sentences in our dataset, and whether players were able to pick up on these differences. For this experiment we found surprisingly different trends across different genres.
Figure \ref{fig:sentence_stats} shows the results for named entities, where novel named entities are ones which occurred in the current sentence but not in any previous sentences.
On News and Speeches, the generated sentences contained fewer named entities than the human-written sentences, while for Stories they had more.
In the News domain, the sentences selected by players tended to have about the same number of entities as the ones not selected while in both Speeches and Stories, the sentences selected by annotators had slightly more named entities.

\section{Discussion and Future Work}
\label{sec:discussion}
In this paper, we demonstrate the viability of the boundary detection task as a framework for soliciting human evaluation of natural-language generation systems.
We conducted the largest study of generated text detectability to date and, in the process, replicated many previous results in the field, such as the improved performance of bigger models \citep{kaplan2020scaling}, the importance of decoding strategy selection \citep{ippolito-etal-2020-automatic}, and the importance of incentivizing annotators \citep{clark2021all}.

In addition, we have provided new insights into the ways in which humans interact with generated text. We have shown that certain textual genres influence models to make different types of errors, that annotators can improve at the detection task over time, and that certain sentence-level features correlate highly with annotator selection.

We expect these results and the released dataset of generations and annotations to be of broad use to those studying detection. However, there is clearly still much left to be done in this space.
One worthwhile avenue for future work would be to study how well automatic systems perform at the detection tasks, and whether we can predict when generated text will be especially difficult for human annotators to recognize.

Future work can also seek to build off our study by testing a larger set of models, genres, and other experimental conditions (finetuning, topic control, decoding strategy, etc.) as well as testing other frameworks for incentivizing annotators.
It is also worth looking at the cases where continuations do not happen exactly on a boundary between sentences, as this is a core limiting assumption of our work.
We also believe that more investigation is needed into exactly what annotators are thinking when they make their decisions and how we can give annotators the right tools to explain their thought processes.

\section{Acknowledgements}

This research is based upon work supported in part by the DARPA KAIROS Program (contract FA8750-19-2-1004), the DARPA LwLL Program (contract FA8750-19-2-0201), the Office of the Director of National Intelligence (ODNI) via the IARPA HIATUS Program (contract 2022-22072200005), and the NSF (Award 1928631). Approved for Public Release, Distribution Unlimited. The views and conclusions contained herein are those of the authors and should not be interpreted as necessarily representing the official policies, either expressed or implied, of DARPA, ODNI, IARPA, NSF, or the U.S. Government. The U.S. Government is authorized to reproduce and distribute reprints for governmental purposes notwithstanding any copyright annotation therein. 

All human subjects research conducted was approved by the Institutional Review Board (IRB) at the University of Pennsylvania  under protocol number 819967. The IRB determined that the proposal meets eligibility criteria for IRB review exemption authorized by 45 CFR 46.101, category 2. 

We would like to thank Google for their generous support through grants of Google Cloud Platform credits to support our server costs.  We also thank Roblox and Salesforce for their financial support through gifts to our research group at the University of Pennsylvania. 

We would like to thank the members of the lab for their suggestions and feedback. In particular, Alyssa Hwang was very
influential in shaping the current version of this
paper. Her great suggestions made the writing
much clearer and much more understandable.

Finally, we are very grateful to the students of the Fall 2021 course CIS521 / CIT596 ``Artificial Intelligence'' for their participation and invaluable contributions to this study. Without their cooperation, none of this research would have been possible.

\bibliography{anthology, roft}

\appendix
\renewcommand{\thetable}{A\arabic{table}}
\renewcommand{\thefigure}{A\arabic{figure}}

\section{Implementation Details}
Sentence segmentation, named entity recognition, and part-of-speech tagging were performed using the spaCy\footnote{https://spacy.io/} \texttt{en\_core\_web\_lg} model.
All tokenization, generation inference, and fine-tuning was done using the HuggingFace Transformers library. As mentioned in the main text, generations were decoded using nucleus sampling with $p$=0.4 unless otherwise specified. For all generations, a maximum context length of 1024 tokens was used along with a repetition penalty of 1.2. All model inference was performed on a single NVIDIA Tesla T4 GPU accessed via Google Cloud Compute Engine. Generating continuations with all models on all datasets took approximately 48 GPU hours.

The fine-tuning of our GPT-2 XL model was performed on the same cloud based NVIDIA Tesla T4 GPU. For fine-tuning, we took our sampled set of 600,000 recipes and split it up into 500k training, 50k validation, and 50k test examples. We then used the HuggingFace Datasets library\footnote{https://github.com/huggingface/datasets} in conjunction with the Trainer module to fine-tune GPT-2 XL. We made a checkpoint every 10,000 examples and calculated perplexity on both the validation set and training set using the model checkpoint. After each of our 500,000 training examples was seen exactly once, we selected the model with the lowest validation set perplexity across all checkpoints to be our final fine-tuned model. This process took approximately 40 GPU hours, however, due to a tokenization error, this process had to be run twice. In total, fine-tuning our model took 80 GPU hours. The final perplexity of the fine-tuned model on the 50k test examples was 4.781 while the pre-trained GPT-2 XL model was 8.979.

\section{Experiments with Larger Models}
\label{app:gpt3}
We held another round of annotation after the initial round to see if the results we found in our experiments with GPT-2 extrapolated to larger models (in particular the GPT-3 family). In this second round we collected about 2,000 annotations in each of the three textual genres (New York Times, Reddit Stories, and Recipes) and of them, about 75\% of generations were from the GPT-3 "Davinci" model and the other 25\% were from GPT-2 XL. The total number of annotations collected is listed in Table \ref{tab:gpt3_data}.

\begin{figure}[tb]
    \center
    \includegraphics[width=0.9\linewidth]{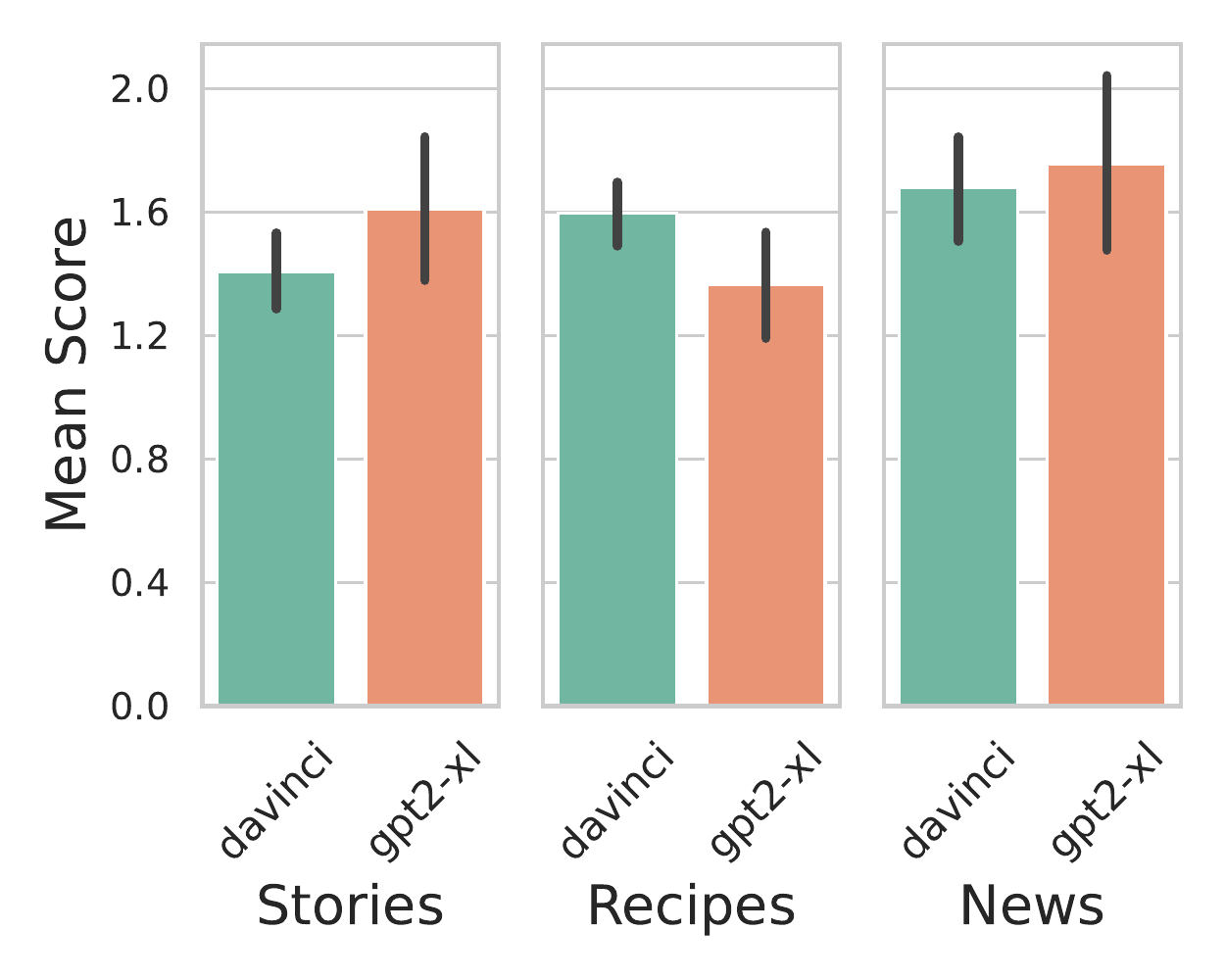}
    \caption{Comparison of mean annotator score between GPT-3 Davinci and GPT-2 XL across three genres of prompt text. Exact numbers can be found in Table \ref{tab:gpt3_data}. We see no statistically significant difference in performance between the two models.}
    \label{fig:gpt3_comparison}
\end{figure}

The results from this extra annotation round are shown in Figure \ref{fig:gpt3_comparison}.
We did not observe a statistically significant difference between GPT-3 and GPT-2 XL with respect to mean annotator score.
There are a number of possible explanations for this, but we mainly attribute it to the nature of the boundary detection task. 
In the more typical binary classification task (i.e., labelling a text passage human-written or machine-generated), annotators are shown the full generation at once and can thus use more context to make their decisions. This is in contrast to the boundary decision task, which only shows one sentence at a time to annotators.
While the boundary detection task has many benefits from an analysis standpoint, it may not be as useful when it comes to comparing models at the highest levels of performance, requiring orders of magnitude more data to draw statistically significant conclusions about model performance.

\begin{table}[tb]
\small
\center
\begin{tabular}{SSSSS} \toprule
    {Dataset} & {Model} & {$p$} & {$n$} & {Mean Score}\\ \midrule
    {Stories} & {GPT-2 XL} & {0.0} & {288} & {1.372 $\pm$ 0.205}\\
    {Stories} & {GPT-2 XL} & {0.4} & {230} & {1.609 $\pm$ 0.237}\\
    {Stories} & {GPT-2 XL} & {1.0} & {253} & {1.743 $\pm$ 0.246}\\\midrule
    {Stories} & {GPT-3 Davinci} & {0.0} & {726} & {1.530 $\pm$ 0.135}\\
    {Stories} & {GPT-3 Davinci} & {0.4} & {776} & {1.406 $\pm$ 0.127}\\
    {Stories} & {GPT-3 Davinci} & {1.0} & {752} & {1.652 $\pm$ 0.134}\\\midrule
    {News} & {GPT-2 XL} & {0.0} & {151} & {1.881 $\pm$ 0.328}\\ 
    {News} & {GPT-2 XL} & {0.4} & {168} & {1.756 $\pm$ 0.295}\\
    {News} & {GPT-2 XL} & {1.0} & {139} & {2.050 $\pm$ 0.358}\\\midrule
    {News} & {GPT-3 Davinci} & {0.0} & {468} & {1.479 $\pm$ 0.171}\\ 
    {News} & {GPT-3 Davinci} & {0.4} & {497} & {1.680 $\pm$ 0.170}\\
    {News} & {GPT-3 Davinci} & {1.0} & {391} & {2.028 $\pm$ 0.203}\\\midrule
    {Recipes} & {GPT-2 XL} & {0.4} & {451} & {1.363 $\pm$ 0.169}\\\midrule
    {Recipes} & {GPT-3 Davinci} & {0.4} & {1,311} & {1.596 $\pm$ 0.103}\\
    \bottomrule
\end{tabular}
\caption{Statistics for our extra second round of annotations. In this round, all generations were either from GPT2-XL or GPT-3 Davinci, in approximately a one to three ratio. Intervals listed are $\alpha=0.95$ confidence.}
\label{tab:gpt3_data}
\end{table}

\section{Dataset Filtering}
\subsection{Quality Control of Generations}
\label{app:gen_filter}
One of the more unexpectedly difficult aspects of this project was ensuring that generations did not have obvious flaws or artifacts that would make the task trivial for annotators. For starters, sentence segmentation with NLTK or spaCy \texttt{en\_core\_web\_sm} resulted in many sentence segmentation errors. Ending quotation marks were tokenized as full sentences and prefixes like Fr. were treated as sentence boundaries. We fixed this, in large part, by switching to the \texttt{en\_core\_web\_trf} model but this solution is not perfect. We look forward to more accurate resources for sentence segmentation in the near future. 

In addition to sentence segmentation, we had difficulty ensuring that sentences generated were complete sentences. Often times generations for News would degenerate into stock ticker readings or lists of addresses or contributors, which are not fun to read for players and are not particularly interesting from a research perspective. In order to solve this, we rejected any generation that did not contain at least one verb in every sentence. We determined part of speech using the same spaCy \texttt{en\_core\_web\_trf} model and looked specifically for the ``VERB'' and ``AUX'' part of speech tags in each sentence.

On top of this filtration, we also encountered many instances where models would generate offensive or unsafe content. In order to filter out this unsafe content we queried the OpenAI API's content filtering endpoint\footnote{https://beta.openai.com/docs/engines/content-filter}. This uses a GPT-3 based unsafe content detection model to label a given set of text as unsafe with a certain confidence. We discarded any generation that was rated as unsafe with over 35.5\% confidence. 

We understand that there may be concerns with our use of quality control measures given that our study directly compares different models. However, we believe that the measures were necessary to preserve the integrity and usefulness of the experiments

\subsection{Filtering Player Annotations}
\label{app:filtering}
\begin{figure}[tb]
    \centering
    \includegraphics[scale=0.4]{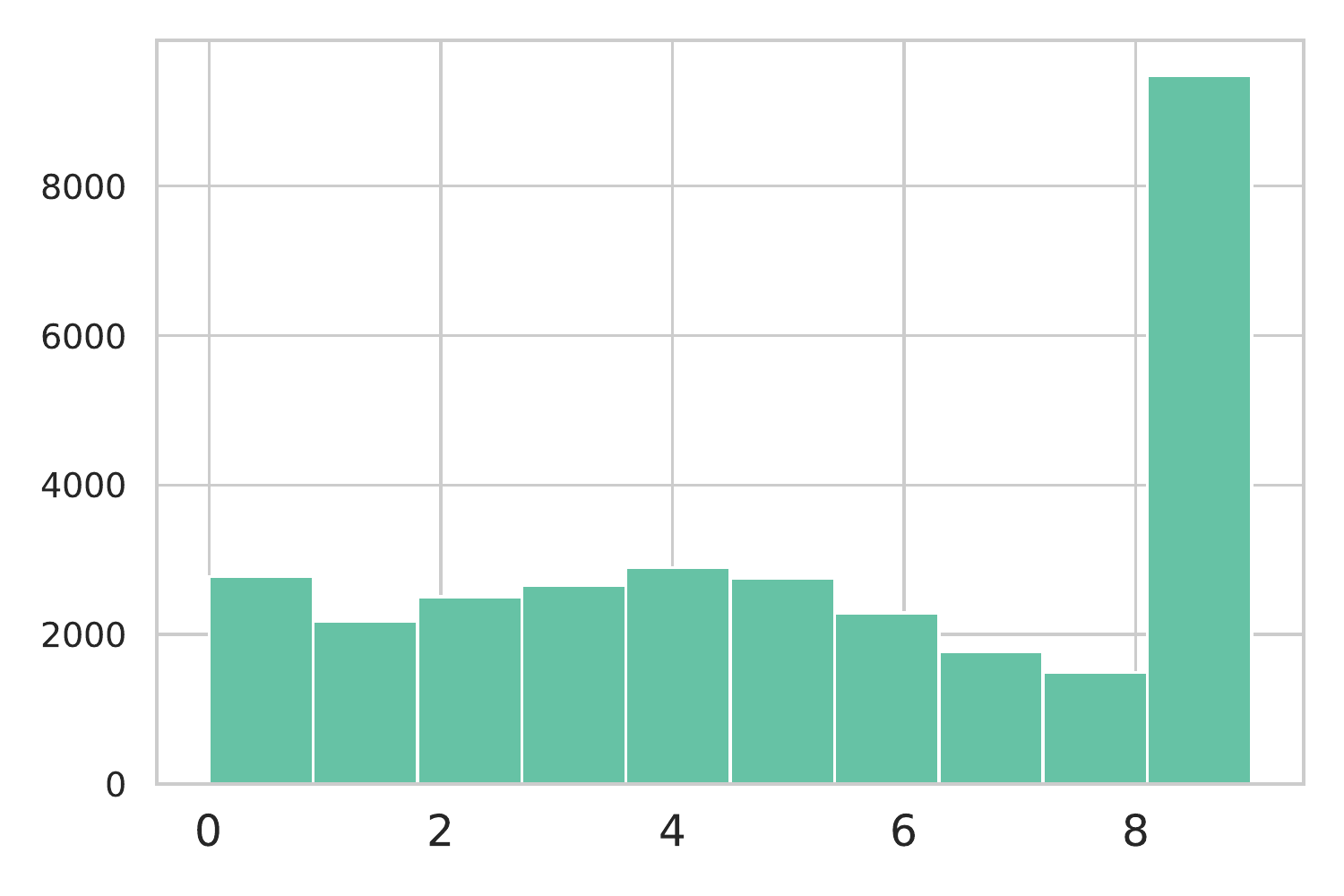}
    \caption{Histogram of predicted boundary index \textit{before filtering annotations}. We can see that a majority of all annotations are labeled as being entirely human written.}
    \label{fig:histogram}
\end{figure}

Over the course of the experiment, we noticed our players had a tendency to gradually start guessing the same boundary sentence multiple times in a row. The typical boundary sentence of choice for this behavior was one sentence after the last (i.e. to annotate the passage as being all human-written). Figure \ref{fig:histogram} shows the observed histogram of predicted boundary indices before filtering. 

\begin{figure}[tb]
    \centering
    \includegraphics[scale=0.35]{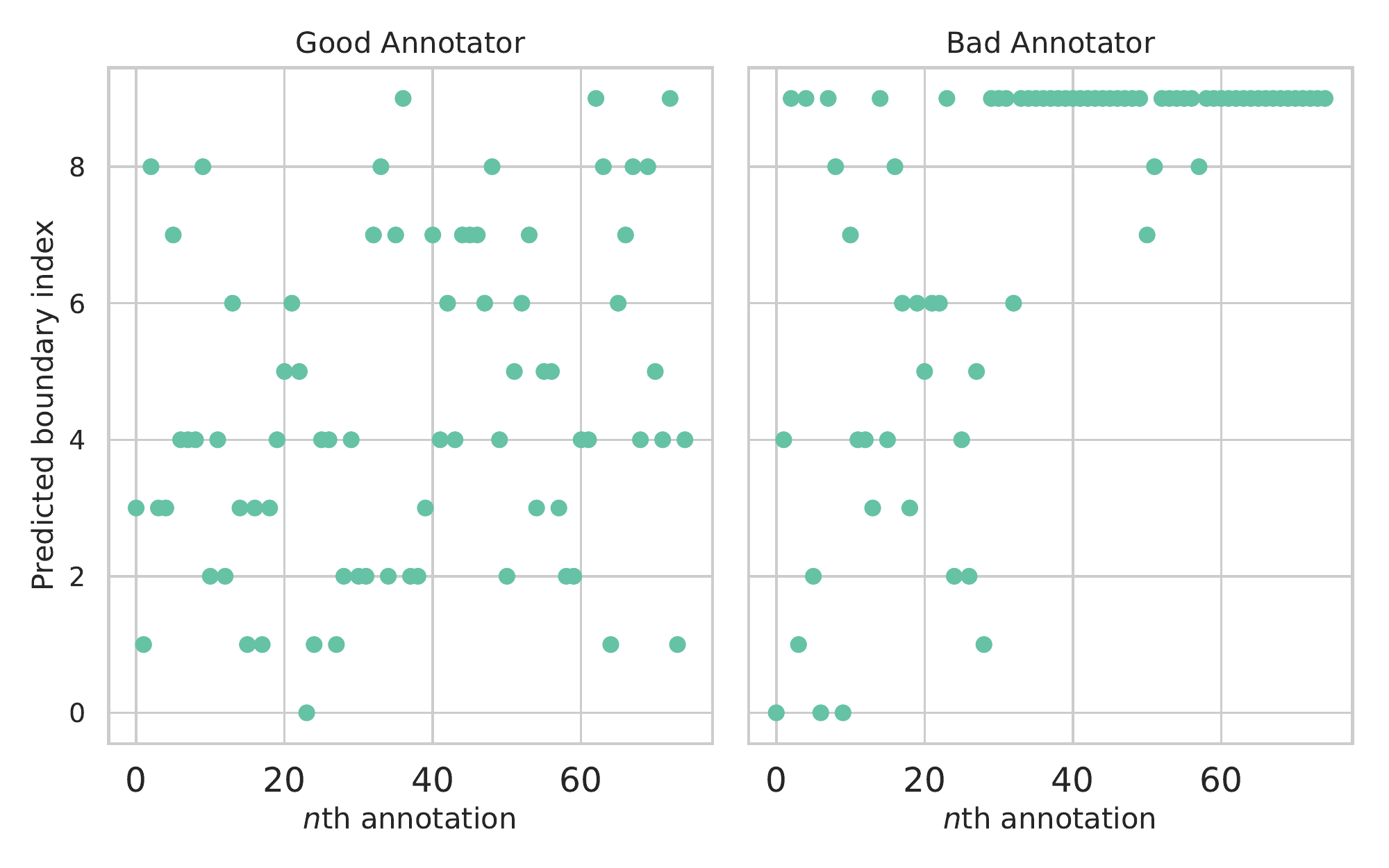}
    \caption{Difference between an honest player (5409) and a dishonest player (5411). We dealt with this by removing any stretch of 5 or more consecutive identical annotations.}
    \label{fig:players}
\end{figure}

To our knowledge, the reason why our players developed this tendency is two-fold. First, if a player guesses all sentences as human-written they do not have to input a reason as to why they made their selection, thus speeding up annotation. Second, due to the nature of our points system, later sentences give more points in expectation than earlier ones. Thus, players that realized this strategy midway through their task were able to more efficiently obtain points by picking the same sentence over and over again. We show an example of this phenomenon in Figure \ref{fig:players}.

In order to identify these rogue players (but not remove the annotations they did before they realized this strategy) we decided to filter out all spans of 5 or more annotations where the player guessed the same index every time. This resulted in us filtering out 3,694 annotations total, with 61 of our 241 players having at least one annotation removed.

\section{Values from Bar Graphs}
\label{app:graph_values}
Tables \ref{tab:top_p}, \ref{tab:size}, and \ref{tab:scores_per_domain} contain the exact values, $\alpha=0.95$ confidence intervals, and $n$-counts for the bar charts in the main paper.

\begin{table}[tb]
\small
\center
\begin{tabular}{SSSS} \toprule
    {Dataset} & {$p$} & {$n$} & {Mean Score}\\ \midrule
    {Stories} & {0.0} & {469} & {1.484 $\pm$ 0.164} \\
    {Stories} & {0.4} & {468} & {1.645 $\pm$ 0.168} \\
    {Stories} & {1.0} & {444} & {2.504 $\pm$ 0.187}  \\\midrule
    {News} & {0.0} & {1,360} & {1.680 $\pm$ 0.102}  \\ 
    {News} & {0.4} & {1,197} & {1.793 $\pm$ 0.109}   \\
    {News} & {1.0} & {1,270} & {2.196 $\pm$ 0.113} \\\bottomrule
\end{tabular}
\caption{Comparison of Generation Performance of GPT2-XL across different values of $p$ and datasets. Interval is $\alpha=0.95$ confidence.}
\label{tab:top_p}
\end{table}

\begin{table}[tb]
\small
\center
\begin{tabular}{SSSSS} \toprule
    {Dataset} & {Model} & {$n$} & {Mean Score}\\ \midrule
    {Stories} & {GPT2} & {2,411} & {2.263 $\pm$ 0.085} \\
    {Stories} & {GPT2-XL} & {613} & {1.645 $\pm$ 0.168} \\
    \midrule
    {Recipes} & {GPT2-XL} & {1,811} & {2.004 $\pm$ 0.098} \\
    {Recipes} & {GPT2-XL (FT)} & {5,157} & {2.184 $\pm$ 0.058} \\
    \midrule
    {Speeches} & {CTRL} & {1,632} & {2.166 $\pm$ 0.099} \\
    {Speeches} & {CTRL-Politics} & {2,620} & {2.174 $\pm$ 0.079} \\
    \bottomrule
\end{tabular}
\caption{Comparison of three different factors (model size, finetuning, and control code) across our three datasets. We see that size (top) has a large effect on human performance while finetuning (middle) and control code usage (bottom) have minimal effect.}
\label{tab:size}
\end{table}

\begin{table}[tb]
\small
\begin{center}
\begin{tabular}{SSSS} \toprule
    {Dataset} & {$p$} & {$n$} & {Mean Score}\\ \midrule
    {Baseline} & {n/a} & {192} & {2.755 $\pm$ 0.307}\\\midrule
    {News} & {0.4} & {1,197} & {1.793$\pm$0.109} \\
    {Stories} & {0.4} & {468} & {1.645$\pm$0.168} \\
    {Speeches*} & {0.4} & {4,252} & {2.171$\pm$0.062}\\
    {Recipes} & {0.4} & {1,811} & {2.004$\pm$0.098} \\
    \bottomrule
\end{tabular}
\caption{The mean scores for each domain on annotations involving XL-sized models for $p = 0.4$. Asterisk denotes generation by CTRL. Interval is $\alpha=0.95$ confidence.}
\label{tab:scores_per_domain}
\end{center}
\end{table}

\section{Metric Correlations}
\label{app:metric}
In Table \ref{tab:correlations_with_other_metrics} we report the correlation between mean score and other sensible metrics. We see that mean score is strongly positively correlated with both perfect guess accuracy and correct side of boundary. Mean score is only weakly correlated with distance after boundary due to the harsh scaling of points; only guesses within five sentences to the right of the boundary receive any points. While imperfect, this harsh scaling is by design, as without it later sentences will give significantly more points in expectation.

\begin{table}[tb]
    \centering
    \small
    \begin{tabular}{c|c}
    \toprule
    Metric & $\rho$ \\
    \midrule
    (a) Correct side of boundary &  0.74 \\
    (b) Perfect guess & 0.88 \\
    (c) Distance after boundary & 0.31  \\
    \bottomrule
    \end{tabular}
    \caption{Spearman's rank correlation between average points per user and several other possible metrics: (a) the fraction of times the user correctly guessed on or after the boundary; (b) the fraction of times the user guessed exactly on the boundary; and (c) the average number of sentences after the boundary of the user's guess (giving new score for guesses before the boundary).}
    \label{tab:correlations_with_other_metrics}
\end{table}

\begin{table*}[tb]
    \centering
    \small
    \begin{tabular}{l|p{30em}}
    \toprule
    Reason & Description \\
    \midrule
    grammar &  is not grammatical \\
    repetition & substantially repeats previous text or itself \\
    irrelevant & is irrelevant or unrelated to the previous sentences \\
    contradicts\_sentence & contradicts the previous sentences \\
    contradicts\_knowledge & contradicts your understanding of the people, events, or concepts involved \\
    common\_sense & contains common-sense or basic logical errors \\
    coreference & mixes up characters' names or other attributes \\
    generic & contains language that is generic or uninteresting \\
    \midrule
    other & $\whitepointerright$Bacon is not sauted \\
     & $\whitepointerright$Mr. vs President Clinton \\
     & $\whitepointerright$navel and sternum seem like very unusual word choices \\
     & $\whitepointerright$It's unlikely that President Nixon will be quoting a one-month old report when he talks about progress made to date	 \\
     & $\whitepointerright$lemon, zest of some things dont sound right? 34 cups of splenda and 14 cups of vinegar? \\
     & $\whitepointerright$doesn't rhyme like rest \\
     & $\whitepointerright$Grammar substantially improves from the previous sentences \\
    \bottomrule
    \end{tabular}
    \caption{\textbf{(top)} The possible reasons players could select for why text was machine generated, and \textbf{(bottom)} several examples of custom reasons players wrote.}
    \label{tab:reasons_text}
\end{table*}

\begin{table*}[tb]
\center
\small
\begin{tabular}{l|rrrrr}
\toprule
Class & \# Participants & \# Annotations & Avg Ann / Part & Avg Score / Part & Avg Time (s)\\
\midrule
Group A & 141 & 6,527 & 46 & 1.966 & {5.651} \\
Group B & 102 & 15,119 & 148 & 2.134 & {6.443} \\
\midrule
Overall & 241 & 21,646 & 90 & 2.083 & {6.338} \\
\bottomrule
\end{tabular}
\caption{
Statistics on the participants and annotations included in our study. ``Avg Ann / Part'' is the average number of annotations per participating student,  while ``Avg Score / Part'' is the average score. ``Avg Time'' is the average time it took a participant to read one sentence. Standard error is shown.
}
\label{tab:particpants}
\end{table*}

\begin{table*}[tb]
    \centering
    \small
    \begin{tabular}{l|l}
    \toprule
    Question & Response Type \\
    \midrule
    What did you (or what are you planning to) major/minor in? & Free Text \\\midrule
    Are you a native English speaker? & Yes/No \\\midrule
    How often do you consult a recipe when preparing food? & Daily (5) \\
    & Once to a few times per week (4) \\
    & Once to a few times per month (3) \\
    & Once to a few times per year (2) \\
    & Never (1) \\\midrule
    How often do you read news from credible news publishers & Daily (5) \\
    (Wall Street Journal, New York Times, etc.)? & Once to a few times per week (4) \\
    & Once to a few times per month (3) \\
    & Once to a few times per year (2) \\
    & Never (1) \\\midrule
    How often do you read fiction on the internet & Daily (5) \\
    (fan fiction, creative writing sub-reddits, ebooks, etc.)? & Once to a few times per week (4) \\
    & Once to a few times per month (3) \\
    & Once to a few times per year (2) \\
    & Never (1) \\\midrule
    What is your familiarity with GPT-2 and GPT-3? & I’ve used them before (OpenAI API, HuggingFace, etc.) (4) \\ 
    & I’ve been excitedly following them. (3) \\
    & I've read about them in the news or a blog post. (2) \\
    & I've never heard of them. (1) \\\midrule
    Did you read the RoFT Guide before you tried the game? & Yes/No \\\midrule
    Do you agree for the data being collected on this form & Yes/No \\
    along with any annotations you make to be used & \\
    in an anonymized, aggregated way for research on participants' ability & \\
    to detect machine-generated text? Your answer on this question & \\ 
    will not affect your grade. & \\
    \bottomrule
    \end{tabular}
    \caption{The text of the exit survey questions given to players after completing their annotations}
    \label{tab:survey_questions}
\end{table*}

\begin{figure*}[tb]
    \centering
    \includegraphics[width=0.9\linewidth]{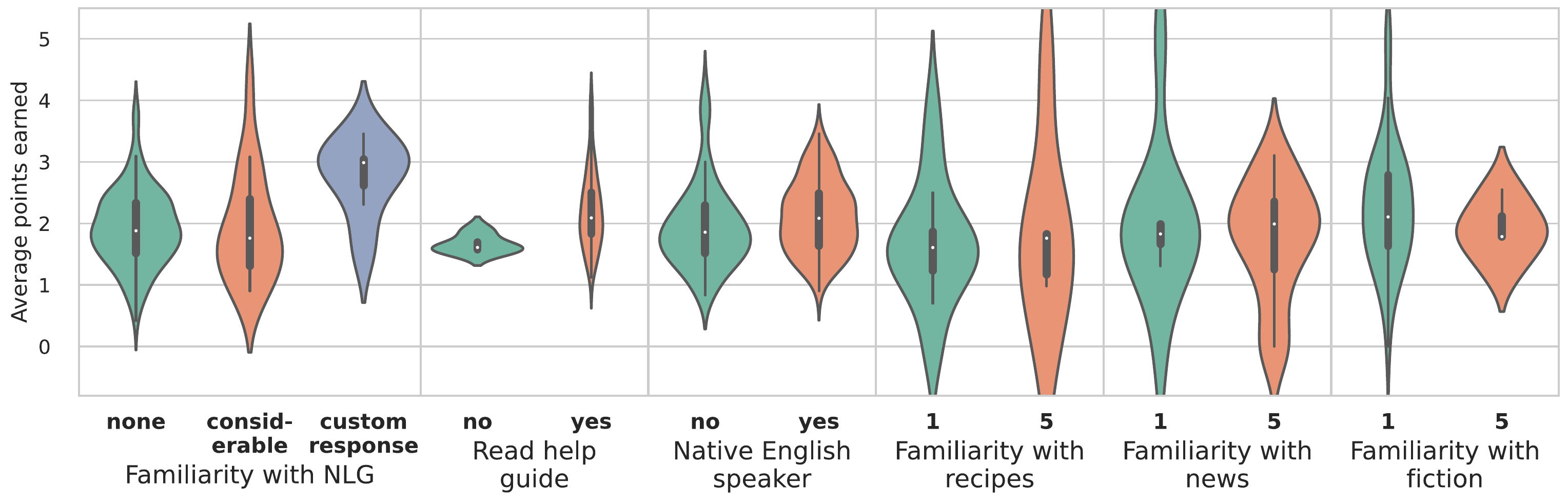}
    \caption{Violin plots showing results of our mandatory exit survey. A violin plot is a box plot that also provides a density estimation. Results shown are filtered to only include players who did at least 20 rounds. Reading the help guide, being a native English speaker, and providing a custom response for familiarity with NLG all correlate very slightly with a higher mean score.
    }
    \label{fig:survey_results_full}
\end{figure*}

\section{Exit Survey Results}
All participating players filled out an exit survey after completing their annotations.
The questions on this survey are listed in full in Table \ref{tab:survey_questions} and selected results are listed in Figure \ref{fig:survey_results_full}. As part of this exit survey, annotators were explicitly asked for their consent to have their data used in this project.
Among the participants who agreed to have their data included, data was collected and fully anonymized to the best of our abilities. We removed email addresses, usernames, and other identifiable information from the dataset file as well as made sure to only ask impersonal and generic survey questions. 

We found that the most impactful feature for predicting annotator skill was whether or not they read our provided help guide. Interestingly, we did not find statistically significant differences in points earned between those self-reported as native English speakers and those who did not. Nor did we find differences between those who reported familiarity with a genre (recipes, fiction, news), and those who did not. 

Finally, while there was no difference between participants who reported they had never heard of GPT-2/3 and those who reported having considerable familiarity with them, interestingly, participants who answered ``other'' and wrote custom responses did tend to perform better at the task. It is worth noting here that the limitations of self-reported qualities on exit surveys are well documented and that, while we did not find any significant correlations, that is not to say that there are none.

\end{document}